\documentclass[manuscript,screen]{acmart}
\AtBeginDocument{%
  }

\begin{document}

\title{Timeseries Foundation Models for Mobility: A Benchmark Comparison with Traditional and Deep Learning Models}

\author{Anita Graser}
\email{anita.graser@ait.ac.at}
\orcid{0000-0001-5361-2885}
\affiliation{%
  \institution{AIT Austrian Institute of Technology}
  \city{Vienna}
  \country{Austria}
}

\renewcommand{\shortauthors}{Graser}
\renewcommand{\shorttitle}{Timeseries Foundation Models for Mobility}

\begin{abstract}
Crowd and flow predictions have been extensively studied in mobility data science. Traditional forecasting methods have relied on statistical models such as ARIMA, later supplemented by deep learning approaches like ST-ResNet. More recently, foundation models for time series forecasting, such as TimeGPT, Chronos, and LagLlama, have emerged. A key advantage of these models is their ability to generate zero-shot predictions, allowing them to be applied directly to new tasks without retraining.
This study evaluates the performance of TimeGPT compared to traditional approaches for predicting city-wide mobility timeseries using two bike-sharing datasets from New York City and Vienna, Austria. Model performance is assessed across short (1-hour), medium (12-hour), and long-term (24-hour) forecasting horizons. The results highlight the potential of foundation models for mobility forecasting while also identifying limitations of our experiments.
\end{abstract}

\maketitle

\section{Introduction}

Accurate mobility forecasting is essential for urban transport operations and planning. Traditionally, time series forecasting has relied on statistical models such as AutoRegressive Integrated Moving Average (ARIMA) and Seasonal ARIMA (SARIMA), along with machine learning and deep learning architectures.

Recent advances in time series foundation models, such as Chronos~\cite{ansari2024chronoslearninglanguagetime}, Lag-Llama~\cite{rasul2024lagllamafoundationmodelsprobabilistic}, and TimeGPT~\cite{garza2024timegpt1}, have demonstrated the potential of large-scale pre-trained models for forecasting. These foundation models leverage knowledge across diverse datasets, providing zero-shot and few-shot forecasting capabilities. However, their effectiveness in mobility prediction remains underexplored.

This study benchmarks TimeGPT against classical and deep learning models for predicting bike-sharing flow and bicycle availability in urban environments. The primary objectives are to: (1) gain intuition about TimeGPT’s ability to predict mobility time series, (2) compare its performance with classical machine learning (ML) and deep learning (DL) models, and (3) analyze how different forecasting horizons impact predictive accuracy.

\section{Related Work}

Mobility forecasting has been widely studied using both traditional and deep learning approaches. To assess developments in flow forecasting over the past decade, we reviewed openly accessible papers citing the seminal ST-ResNet paper~\cite{Zhang_Zheng_Qi_2017} and referencing the BikeNYC dataset using a targeted Google Scholar search query (\url{https://scholar.google.at/scholar?hl=en&as_sdt=2005&sciodt=0%2C5&cites=10585174952970430867&scipsc=1&q=bikenyc&btnG=}). 
Papers that provide comparable BikeNYC experiments include \cite{Mourad2019, xiong2022clstan}.
Their results are included in the following comparisons.
Multiple other papers mention BikeNYC but use different datasets from the NYC bicycle sharing system \cite{du2020traffic, ye2021coupled, jiang2021dl}
or don't provide enough information to ensure that the experiments are comparable \cite{parvathi2024enhancing}.

Recent literature also introduces explicitly spatiotemporal foundation models such as UrbanGPT~\cite{li2024urbangptspatiotemporallargelanguage}, UniST~\cite{Yuan_2024}, and UrbanDiT~\cite{yuan2024urbanditfoundationmodelopenworld}. However, to our knowledge, none have publicly released their model weights, limiting reproducibility and further research.

\section{Experimental Setup}

Our experiments build upon the ST-ResNet framework~\cite{Zhang_Zheng_Qi_2017}. Since the original code repository (\url{https://github.com/lucktroy/DeepST/tree/master/scripts/papers/AAAI17}) is no longer available, we leveraged a publicly available implementation \url{https://github.com/topazape/ST-ResNet}. The complete experimental setup is available at \url{https://github.com/anitagraser/ST-ResNet/blob/main/experiment.ipynb}.

This study uses two datasets: BikeNYC and BikeVIE, summarized in Table~\ref{tab:datasets}. BikeNYC has been extensively studied, while BikeVIE provides a less-explored dataset from Vienna, Austria for evaluating generalization. The BikeVIE timeseries were preprocesssed by resampling  the number of available bicycles to the maximum hourly value, with a cutoff in mid-August to avoid a longer data collection gap as well as the less busy autumn and winter period.

\begin{table}[h]
  \caption{Datasets}
  \label{tab:datasets}
\centering

\begin{tabular}{l p{6cm} p{6cm}}
\toprule
 & BikeNYC & BikeVIE \\
\midrule
Description: & Hourly bike flows in New York City  & Hourly bike-sharing station data from Vienna \\
Source: & \url{https://1drv.ms/f/s!Akh6N7xv3uVmhOhCtwaiDRy5oDVIug} & \url{https://www.opendataportal.at/katalog/de/dataset/sharedmobility-ai-tensorflow-datasets}\\
No. timeseries: & 16x8=128 grid cells & 120 stations \\
Time span: & 2014-04-01 to 2014-09-30 & 2019-05-07 to 2019-12-31 (cut off at 2019-08-15) \\
\bottomrule
\end{tabular}
\end{table}

Benchmark models include:
\begin{itemize}
\item \textbf{Baseline models}: AutoARIMA, Seasonal Naive, Historical Average 
\item \textbf{Foundation model}: TimeGPT 
\end{itemize}

\begin{figure}[h]
  \centering
  \includegraphics[width=\linewidth]{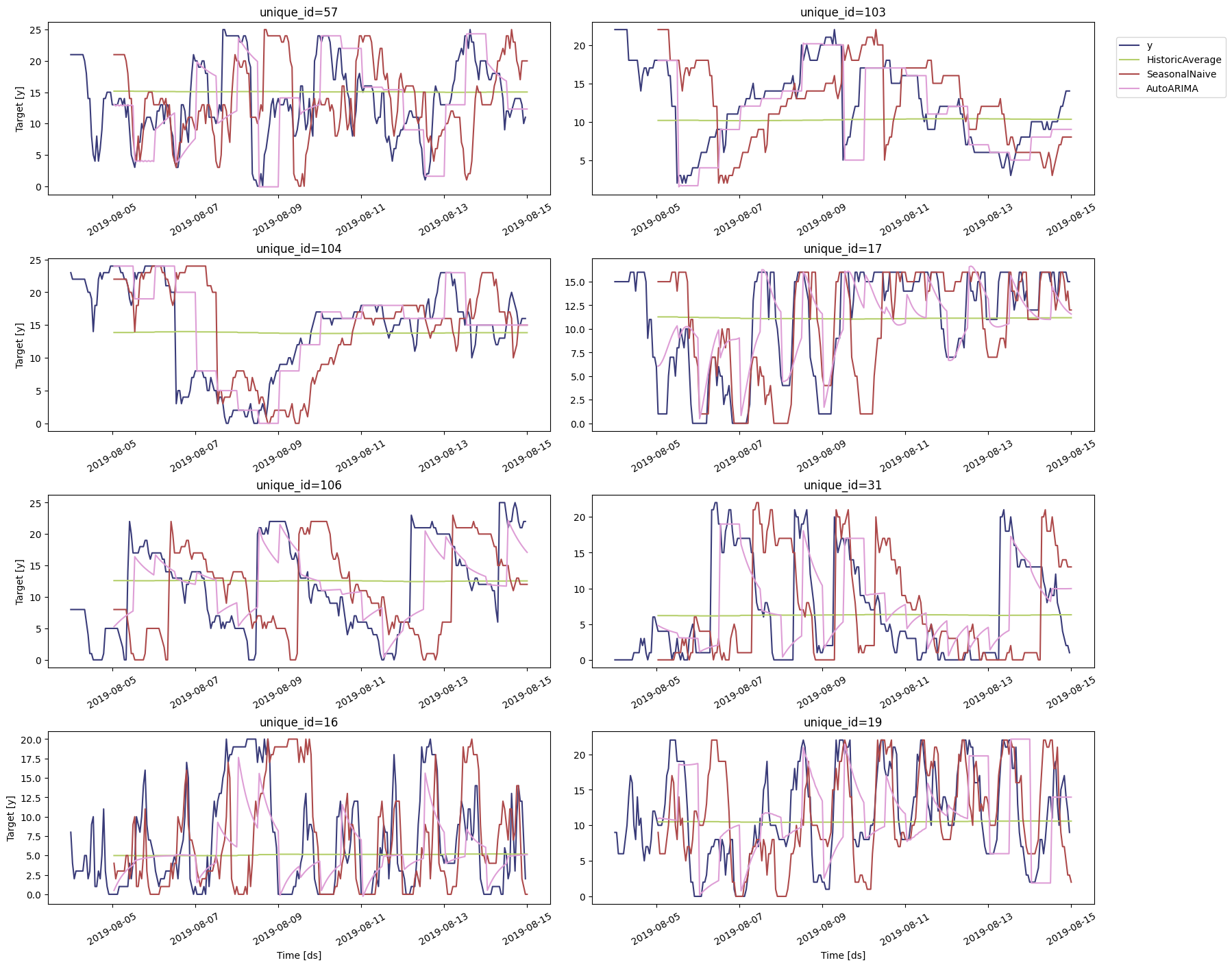}
  \caption{Baseline 12-hour forecast examples for BikeVIE.}
  \label{fig:baseline-h12}
\end{figure}

\begin{figure}[h]
  \centering
  \includegraphics[width=\linewidth]{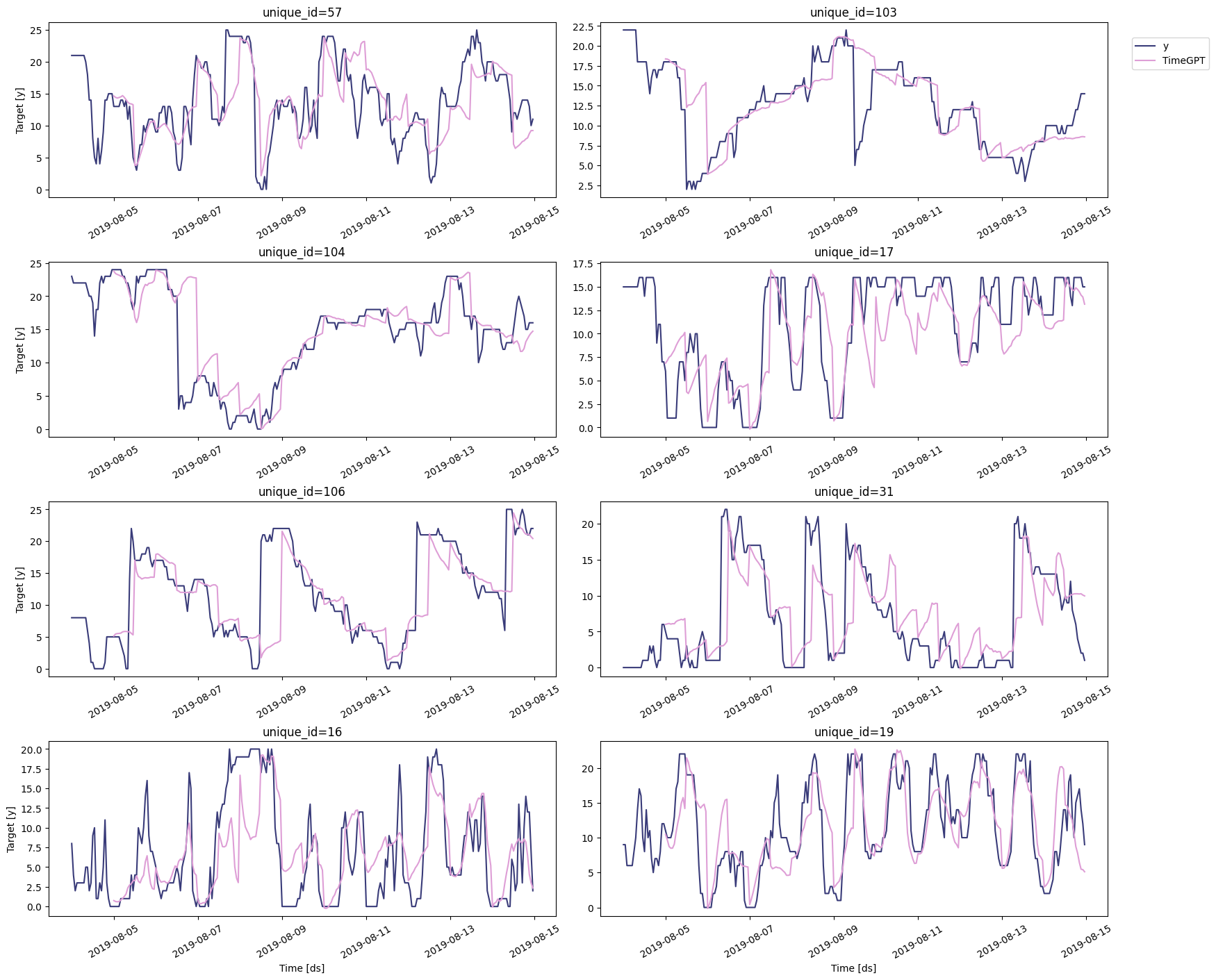}
  \caption{TimeGPT 12-hour forecast examples for BikeVIE.}
  \label{fig:timegpt-h12}
\end{figure}

Example forecasting results are provided in Figure~\ref{fig:baseline-h12} and Figure~\ref{fig:timegpt-h12} for the baseline models and TimeGPT, respectively.

Performance is evaluated using Root Mean Square Error (RMSE) for 1-hour, 12-hour, and 24-hour forecasts, with the last ten days used for backtesting via a rolling window approach.

\newpage
\section{Results and Discussion}

In this section, we present the results obtained for the BikeNYC and BikeVIE datasets. For BikeNYC, we compare TimeGPT to model performance results found in the literature as well as to our three baseline models. For BikeVIE, we rely solely on our three baseline models since, to the best of our knowledge, there are no comparable results in the literature.

\subsection{BikeNYC Results}

\begin{figure}[h]
  \centering
  \includegraphics[width=\linewidth]{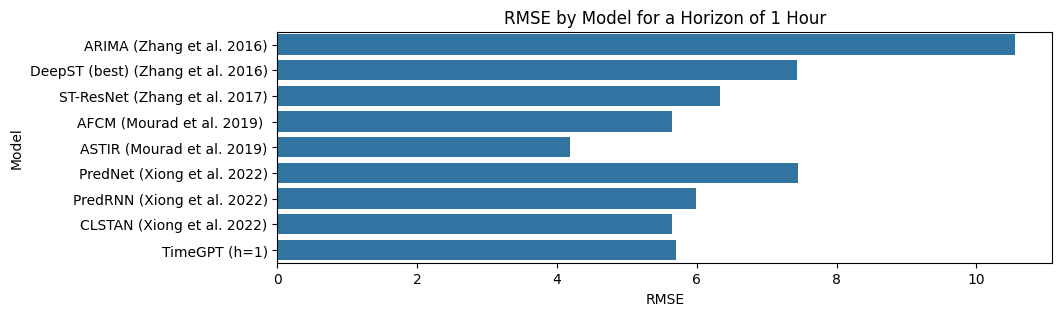}
  \caption{BikeNYC results. Comparing model results for a horizon of 1 hour.}
  \label{fig:bikenyc1}
\end{figure}

Table~\ref{tab:bikenyc} summarizes the results for the BikeNYC dataset. TimeGPT outperforms ARIMA, DeepST, ST-ResNet, PredNet, and PredRNN for 1-hour forecasts (Figure~\ref{fig:bikenyc1}). Only ASTIR~\cite{Mourad2019} reports notably better results.

\begin{figure}[h]
  \centering
  \includegraphics[width=\linewidth]{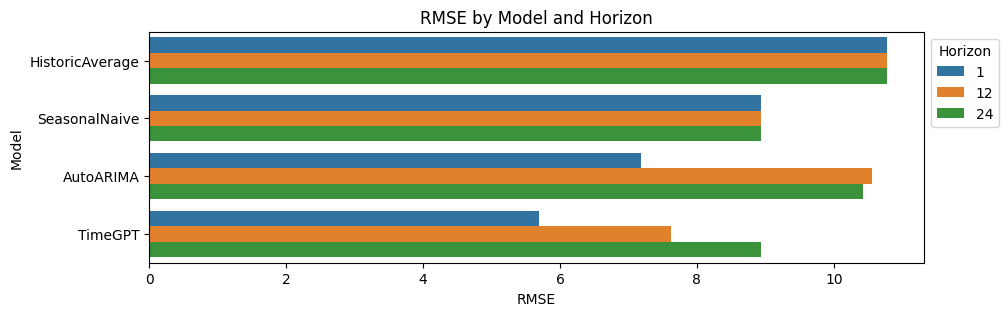}
  \caption{BikeNYC results. Comparing model results for different forecast horizons.}
  \label{fig:bikenyc124}
\end{figure}

For longer horizons, TimeGPT remains competitive to the baselines but converges to the SeasonalNaive performance at 24 hours (Figure~\ref{fig:bikenyc124}). Notably, SeasonalNaive outperforms AutoARIMA for 12-hour and 24-hour forecasts, suggesting strong seasonal effects.

\begin{table}[h]
\centering
  \caption{BikeNYC results}
  \label{tab:bikenyc}

\begin{tabular}{l c r}
\toprule
Model & Horizon & RMSE \\
\midrule
ARIMA \cite{zhang2016dnn} & 1 & 10.56 \\
SARIMA \cite{zhang2016dnn} & 1 & 10.07 \\
VAR \cite{zhang2016dnn} & 1 & 9.92 \\
DeepST \cite{zhang2016dnn} & 1 & 7.43 \\
ST-ResNet \cite{Zhang_Zheng_Qi_2017} & 1 & 6.33 \\
AFCM \cite{Mourad2019} & 1 & 5.64 \\
\textbf{ASTIR} \cite{Mourad2019} & \textbf{1} & \textbf{4.18} \\
PredNet \cite{xiong2022clstan} & 1 & 7.45 \\
PredRNN \cite{xiong2022clstan} & 1 & 5.99 \\
CLSTAN \cite{xiong2022clstan} & 1 & 5.65 \\
\midrule
HistoricAverage & 1/12/24 & 10.78 \\
SeasonalNaive & 1/12/24 & 8.93 \\
AutoARIMA & 1 & 7.18 \\
AutoARIMA & 12 & 10.55 \\
AutoARIMA & 24 & 10.42 \\
TimeGPT & 1 & 5.70 \\
TimeGPT & 12 & 7.62 \\
TimeGPT & 24 & 8.93 \\
\bottomrule
\end{tabular}

\end{table}

\newpage
\subsection{\textbf{BikeVIE Results}}

Table~\ref{tab:bikevie} summarizes the results for the BikeVIE dataset.  
TimeGPT struggles with longer horizons, with RMSE more than doubling from 1-hour to 12-hour forecasts. While TimeGPT marginally outperforms AutoARIMA at 12 and 24 hours, the results indicate significant forecast uncertainty. The poor performance of SeasonalNaive, even compared to the simple HistoricAverage, suggests the presence of irregular station restocking patterns that affect predictability.

\begin{figure}[h]
  \centering
  \includegraphics[width=\linewidth]{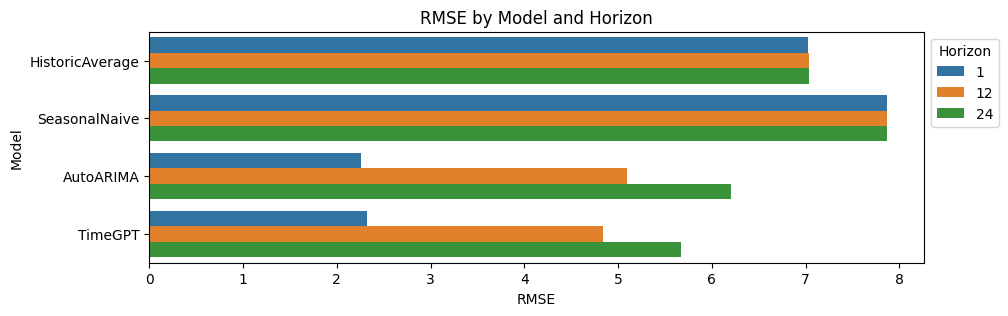}
  \caption{BikeVIE results. Comparing model results for different forecast horizons.}
  \label{fig:bikevie}
\end{figure}

\begin{table}[h]
  \caption{BikeVIE results}
  \label{tab:bikevie}
\centering

\begin{tabular}{l c r}
\toprule
Model & Horizon & RMSE \\
\midrule
HistoricAverage & 1/12/24  & 7.04 \\
SeasonalNaive & 1/12/24 & 7.87  \\
\textbf{AutoARIMA} & \textbf{1}  & \textbf{2.26} \\
AutoARIMA & 12  & 5.10 \\
AutoARIMA & 24  & 6.20 \\
TimeGPT & 1  & 2.32 \\
TimeGPT & 12  & 4.84 \\
TimeGPT & 24  & 5.67 \\
\bottomrule
\end{tabular}
\end{table}

\section{Conclusion and Future Work}

These results indicate that time series foundation models like TimeGPT hold promise for mobility forecasting. However, a key limitation is the potential exposure of foundation models to datasets used in evaluation, raising concerns about true generalization. Future studies should evaluate foundation models on previously unpublished datasets.

Since deep learning models benefit from additional training data, it remains likely that domain-specific deep learning models will outperform foundation models when sufficiently trained. However, in data-sparse scenarios, foundation models present a viable forecasting option. 

Future research should furthermore explore the impact of auxiliary information, such as weather and event data, on model predictive performance when incorporated as covariates.

\begin{acks}
This work was funded by the EU’s Horizon Europe research and innovation program under Grant No. 101093051 EMERALDS.

During the preparation of this manuscript, the author used ChatGPT in order to paraphrase and reword the text and to write code to generate figures. After using ChatGPT, the author reviewed and edited the content as needed and takes full responsibility for the manuscript's content.
\end{acks}

\bibliographystyle{ACM-Reference-Format}
\bibliography{sample-base}

\newpage
\section*{Appendix}

\end{document}